\RequirePackage{iftex}
\ifPDFTeX
\else
\RequirePackage[T1]{fontenc}
\fi
\documentclass[letterpaper,journal]{IEEEtran}

\usepackage{amsmath,amsfonts,amssymb}
\usepackage{array}
\usepackage{booktabs}
\usepackage{cite}
\usepackage{graphicx}
\usepackage{microtype}
\usepackage{makecell}
\usepackage{multirow}
\usepackage[table]{xcolor}
\usepackage{textcomp}
\usepackage{url}
\usepackage{xspace}
\ifXeTeX
\usepackage{newunicodechar}
\newunicodechar{’}{'}
\fi
\usepackage{enumitem}
\usepackage{stfloats}
\usepackage{placeins}
\usepackage[capitalise,nameinlink]{cleveref}

\newcommand{\gain}[1]{\textcolor{teal}{\raisebox{0.25ex}{\scalebox{0.7}{$\uparrow$}}\,\textbf{#1}}}

\crefname{figure}{Fig.}{Figs.}
\Crefname{figure}{Fig.}{Figs.}
\crefname{table}{Table}{Tables}
\Crefname{table}{Table}{Tables}

\makeatletter
\newenvironment{attachedtable}{\par\def\@captype{table}}{\par}
\def\@IEEENORMtitlevspace{1.25\baselineskip}
\def\@IEEEMINtitlevspace{1\baselineskip}
\makeatother
\ifPDFTeX
\else

\fi
\begin{document}
\raggedbottom

\title{Can BEV Perception Gracefully Degrade under Sensor Failures?}

\author{Haifa~Zhang, Yijing~Wang, and Zhiqiang~Zuo,~\IEEEmembership{Senior~Member,~IEEE}%
\thanks{All authors are with the Tianjin Key Laboratory of Intelligent Unmanned Swarm Technology and System, School of Electrical and Information Engineering, Tianjin University, Tianjin 300072, China (e-mails: zhanghaifa@tju.edu.cn; yjwang@tju.edu.cn; zqzuo@tju.edu.cn).}%
}
\markboth{Preprint}{Zhang et al.: Can BEV Perception Gracefully Degrade under Sensor Failures?}

\maketitle

\begin{abstract}
Despite the remarkable success of multi-modal bird's-eye view (BEV) perception in autonomous driving, standard fusion models can be highly brittle to sensor corruptions, causing catastrophic performance degradation.
This vulnerability arises when corrupted features contaminate the shared BEV representation and the detector fails to preserve a usable visual fallback.
To this end, we present Grace-BEV, a lightweight framework for robust fusion in Lift-Splat-Shoot (LSS)-based detectors.
Operating in the aligned BEV space, Grace-BEV complements a LiDAR-guided image-BEV expert with a LiDAR-independent visual expert. A TrustGate Router learns a sample-level mixture of their outputs, while the FailSafe Fusion Block recalibrates the routed image representation together with raw LiDAR features.
Furthermore, we devise a Three-Phase Training strategy with Modality Dropout to establish the visual fallback before learning routing and fusion, followed by joint fine-tuning.
Extensive experiments on nuScenes-R and nuScenes-C across three LSS-based detectors show that Grace-BEV maintains robust performance across diverse corruption settings. Under complete LiDAR loss, where the corresponding baseline models exhibit a collapse to 0.0 mAP, Grace-BEV restores performance to as high as 34.7 mAP. It also improves clean accuracy by up to 1.4 mAP points with 1.2M additional parameters per detector. Component and training ablations support the complementary roles of visual initialization, expert routing, and adaptive fusion.
\end{abstract}


\section{Introduction}
\label{sec:intro}

In recent years, bird's-eye view (BEV) perception~\cite{li2022unifying,liu2022bevfusion,liang2022bevfusion,cai2023objectfusion} has emerged as a fundamental representation for autonomous driving, providing a unified interface for 3D object detection and motion planning~\cite{2023uniAD,2023wanglong}. Driven by the complementary strengths of cameras and LiDAR, multi-modal fusion has become the dominant paradigm for high-precision BEV perception~\cite{chen2017multi,vora2020pointpainting,2024wangbevspread}. Current state-of-the-art (SOTA) frameworks achieve remarkable performance by aligning heterogeneous sensor streams and integrating dense semantic cues from cameras with precise geometric information from LiDAR into an aligned BEV space~\cite{zhao2024simplebev,vaswani2017attention,li2025bevformer,yan2023cmt}.

\begin{figure*}[t]
  \centering
  \includegraphics[width=0.92\textwidth]{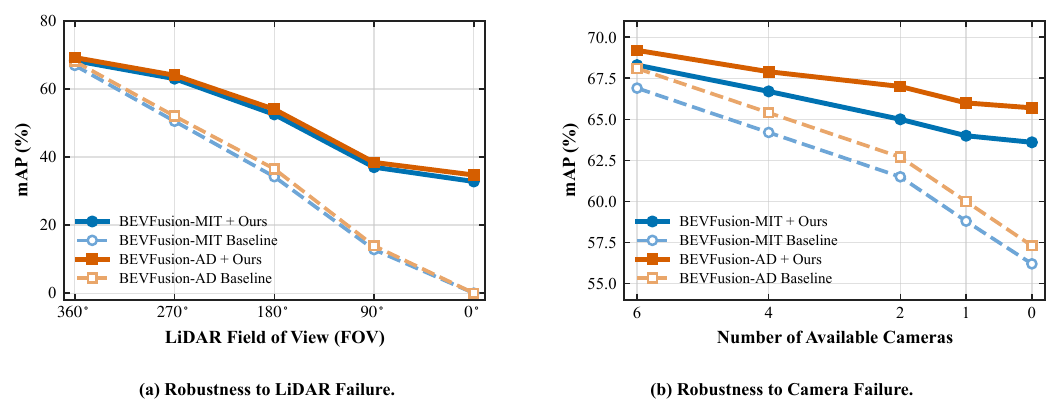}
  \caption{\textbf{Sensor-failure responses.} Grace-BEV (solid) and the corresponding baseline models (dashed), evaluated on BEVFusion-MIT (blue) and BEVFusion-AD (orange), under \textbf{(a)} reduced LiDAR FOV and \textbf{(b)} camera loss.}
  \label{fig:robustness_analysis}
\end{figure*}

However, this strong performance can degrade when the assumption that all modalities remain reliable no longer holds~\cite{chen2024end,BADUE2021113816}. In real-world deployments, sensor streams are often incomplete or corrupted, which poses a fundamental challenge to multi-modal fusion. Static and tightly coupled fusion can be brittle to missing or unreliable modalities. More importantly, once an unreliable modality is fused into the shared BEV representation, its corrupted features can suppress otherwise valid cues from the remaining modality. As illustrated in~\cref{fig:robustness_analysis}, when the primary geometric modality (e.g., LiDAR) is severely degraded, the BEVFusion baselines evaluated in this study~\cite{liu2022bevfusion,liang2022bevfusion} fail to degrade gracefully and instead exhibit an abrupt collapse in detection performance. One contributing factor is modality dominance, where the model over-relies on the geometry-rich modality during fusion and fails to preserve a viable fallback when that modality becomes unreliable~\cite{peng2022balanced,wu2022characterizing}. Recent robust fusion schemes, including MetaBEV, RobBEV, UniBEV, and MoME, already handle missing sensors or unreliable feature interactions~\cite{ge2023metabev,wang2024robbev,wang2024unibev,mome}. Our focus is to preserve a usable visual fallback and adapt its cooperation with LiDAR-related representations within an efficient LSS-based pipeline.

To address this challenge, we present Grace-BEV, a lightweight framework for robust multi-modal fusion in LSS-based BEV perception. We couple the preservation of a LiDAR-independent visual pathway with adaptive feature interactions in the aligned BEV space. Specifically, a TrustGate Router learns a sample-level mixture of LiDAR-guided and LiDAR-independent image-BEV experts from LiDAR feature statistics. The proposed FailSafe Fusion Block (FFB) then dynamically recalibrates the routed image representation together with raw LiDAR features before downstream prediction. Furthermore, we devise a Three-Phase Training strategy with Modality Dropout (MD)~\cite{yan2023cmt,wang2021pointaugmenting} to establish the visual fallback, warm up routing and fusion on frozen expert features, and jointly fine-tune the detector under unreliable inputs. Our main contributions are summarized as follows:
\begin{itemize}[topsep=4pt, partopsep=0pt, parsep=0pt]
\item We study cross-modal contamination and modality dominance in tightly coupled BEV fusion, and organize robust perception around preserving a usable visual fallback and adapting its interaction with LiDAR-related representations.
\item We propose \textbf{Grace-BEV}, a lightweight framework for robust multi-modal fusion. The framework introduces a TrustGate Router to adapt the mixture of the two image-BEV experts and an FFB to recalibrate their routed output together with raw LiDAR features in the shared BEV representation.
\item We further devise a Three-Phase Training strategy that establishes visual capability before learning routing and fusion under MD. Extensive experiments on nuScenes-R and nuScenes-C across three LSS-based detectors show that Grace-BEV restores performance from 0.0 mAP for the corresponding baseline models to as high as 34.7 mAP under complete LiDAR loss, while also improving clean accuracy by up to 1.4 mAP points, achieving a strong trade-off between robustness and efficiency.
\end{itemize}

\section{Related Work}
\label{sec:work}

\subsection{Multi-modal 3D Object Detection}
Existing multi-modal 3D detectors for autonomous driving mainly follow two paradigms for aligning heterogeneous observations.

\textbf{Geometry-based alignment approaches}, such as Lift, Splat, Shoot (LSS)~\cite{philion2020lift} and BEVFusion~\cite{liu2022bevfusion,liang2022bevfusion}, explicitly lift image features into 3D space via depth estimation and project them into a unified BEV grid.
By leveraging explicit spatial priors to align dense image semantics with sparse point clouds, these methods achieve strong performance and high efficiency under normal conditions.
Static fusion strategies such as feature concatenation can be sensitive to corrupted geometric signals and may allow unreliable features to contaminate the shared BEV representation under modality failures.
BEVDepth and BEVStereo improve image-to-BEV construction through depth supervision and temporal stereo~\cite{li2023bevdepth,li2023bevstereo}; BEVHeight++ combines height- and depth-based representations across roadside and ego-vehicle settings~\cite{yang2025bevheight}.

\textbf{Query- and transformer-based methods}, including TransFusion, CMT, and DeepInteraction, model cross-modal interactions through attention mechanisms or learned queries~\cite{li2025bevformer,liu2022petr,wang2022detr3d,chen2023focalformer3d,wang2023unitr,yan2023cmt,bai2022transfusion,yang2022deepinteraction,yang2025deepinteraction++}.
Related approaches align deep modality features or maintain sparse representations~\cite{li2022deepfusion,xie2023sparsefusion,lin2022sparse4d}, while temporal feature learning further supports detection~\cite{zong2023Temporal}.
Other fusion designs combine local and global features in LoGoNet or operate on sparse instances in Fully Sparse Fusion~\cite{li2023logonet,li2024fully}.

Our work is motivated by the observation that LSS-based architectures already align multi-modal features explicitly in BEV space, enabling their interactions to be adapted directly through lightweight expert routing and fusion.

\subsection{Robustness in Multi-modal Perception}
Robust multi-modal perception is critical for autonomous driving, where sensor failures and degradations are unavoidable~\cite{yu2023benchmarking,dong2023benchmarking,kong2023robo3d}.
Existing approaches generally address robustness from two perspectives: training strategies and architectural designs.

\noindent\textbf{Training Strategies.}
A common practice is MD~\cite{wang2021pointaugmenting}, which randomly removes one modality during training to discourage over-reliance on a dominant modality.
We adopt the MD distribution used in CMT: 0.25 camera-only, 0.25 LiDAR-only, and 0.50 multimodal samples~\cite{yan2023cmt}.
Within this distribution, our three-phase schedule establishes the visual pathway before optimizing routing and fusion, followed by joint fine-tuning.

\noindent\textbf{Architectural Designs.}
Beyond training-time regularization, several works redesign fusion architectures to better handle partial or corrupted observations.
MetaBEV selectively aggregates available modalities, while RobBEV addresses unreliable LiDAR--camera interactions through mutual deformable attention and temporal aggregation~\cite{ge2023metabev,wang2024robbev}.
UniBEV studies uniform BEV encoders for missing modalities, and MoME and MEFormer use expert-based designs for sensor-failure robustness~\cite{wang2024unibev,mome,cha2024meformer}.

Knowledge distillation provides a different deployment setting. MonoDistill transfers LiDAR knowledge to a monocular detector~\cite{chong2022monodistill}, whereas CMDFusion transfers image knowledge into a LiDAR semantic-segmentation network that requires no camera input at inference~\cite{cen2023cmdfusion}. In contrast, Grace-BEV retains camera and LiDAR inputs within a 3D detection system and adaptively modulates fusion when either input becomes unavailable or unreliable.

Preserving single-sensor capability has also been investigated in CSMSF for robust depth estimation~\cite{hu2025robustdepth}. In addition, parameter-efficient adaptation has been explored for handling missing modalities in frozen multimodal networks using a small set of learnable parameters~\cite{reza2025missing}.
Our approach focuses on LSS-based BEV pipelines: it establishes a visual fallback, learns its mixture with a LiDAR-guided image representation, and recalibrates the result with raw LiDAR features before a shared detection head.

\begin{figure*}[tp]
  \centering
  \includegraphics[width=0.8\textwidth]{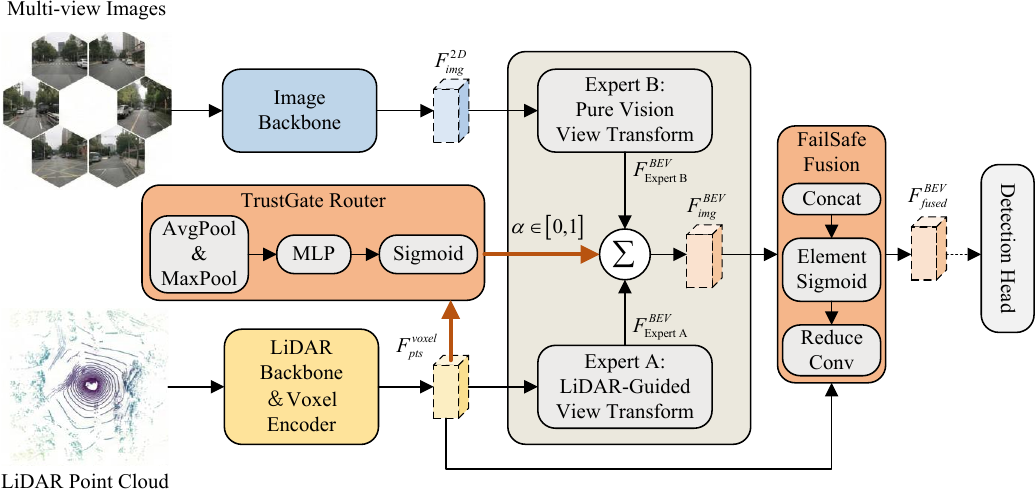}
  \caption{
  \textbf{Overview of the Grace-BEV Framework.}
  The two image-BEV experts use LiDAR-guided projection (Expert A) and LiDAR-independent projection (Expert B), respectively.
  A TrustGate Router predicts a sample-adaptive coefficient $\alpha$ from globally pooled Voxel Encoder statistics.
  Features are then integrated via the FFB, which employs element-wise Sigmoid gating to recalibrate the routed image feature together with the raw LiDAR BEV feature.
  }
  \label{fig:framework}
\end{figure*}

\subsection{Dynamic Routing and Conditional Computation}
Dynamic routing and conditional computation have been explored in prior work to adapt model behavior based on input characteristics, most notably in Mixture-of-Experts (MoE) frameworks~\cite{shazeer2017outrageously,riquelme2021scaling}.
While conventional MoE designs primarily target capacity scaling or computational efficiency~\cite{fedus2022switch}, our work uses routing to adapt the contributions of two aligned image-BEV representations with a scalar mixing coefficient.
Both experts are evaluated, and the router adapts their feature contributions; FFB provides the subsequent spatially varying recalibration.

\section{Problem Analysis}
\label{sec:problem}

While multi-modal BEV detectors generally outperform single-modal baselines, their reliability under modality failure remains a critical vulnerability~\cite{yu2023benchmarking,dong2023benchmarking,kong2023robo3d}.
As evidenced by our empirical stress tests in \cref{fig:robustness_analysis}, the BEVFusion baselines evaluated in this study exhibit a ``cliff-like'' performance drop.
Specifically, as shown in the left panel of \cref{fig:robustness_analysis}, when LiDAR field of view (FOV) is reduced to $0^{\circ}$ (simulating a complete sensor blackout), the mean Average Precision (mAP) of baseline methods (dashed lines) collapses to near-zero, indicating a failure to utilize the remaining valid camera signals.
By comparison, the right panel of \cref{fig:robustness_analysis} shows a milder decline when cameras are lost and LiDAR remains available.

Current attempts to mitigate this fragility include MoE~\cite{mome,cha2024meformer} and query-based interactions~\cite{bai2022transfusion,yang2022deepinteraction,lin2022cat}.
The central issue is adapting a corrupted modality's feature contribution while preserving useful evidence from the remaining modality.
For LSS-based architectures~\cite{philion2020lift}, semantic features from different modalities are explicitly projected and unified into a common BEV coordinate system.
In this unified space, multi-modal features are spatially aligned by design.
This common interface allows the contributions of aligned BEV features to be adapted directly through lightweight routing and fusion.

This insight motivates the design of Grace-BEV.
We propose to leverage the unified BEV space to:
1) learn a sample-level mixture of LiDAR-guided and LiDAR-independent image-BEV experts via a lightweight router;
and 2) dynamically \textit{recalibrate} the routed image representation together with raw LiDAR features via element-wise gating.
These adaptations build on a LiDAR-independent visual pathway; routing and fusion learn how to combine its features with LiDAR-related representations.
This supports graceful degradation with modest architectural overhead, preserving the efficiency inherent to LSS pipelines.

\section{Methodology}
\label{sec:method}

\subsection{Overview of Grace-BEV Framework}
\label{subsec:overview}

Grace-BEV preserves two image-to-BEV pathways within a lightweight MoE design (\cref{fig:framework})~\cite{shazeer2017outrageously,mome}. Given multi-view images $X_{img}$ and LiDAR point clouds $X_{lidar}$, modality-specific backbones extract the initial features~\cite{liu2021swin,yan2018second,zhou2018voxelnet,lang2019pointpillars}.

\noindent\textbf{Dual Experts and Adaptive Fusion.} Expert A uses LiDAR-derived geometric guidance for image-to-BEV projection, while Expert B constructs BEV features from image features through a LiDAR-independent view transform~\cite{philion2020lift,li2023bevdepth}. TrustGate predicts a coefficient $\alpha\in[0,1]$ from LiDAR feature statistics to mix the two expert outputs. FFB then applies element-wise recalibration to the routed image representation and raw LiDAR BEV features.

\noindent\textbf{Applicability through the BEV Interface.} Integration requires compatible spatial and channel dimensions for the expert outputs and a LiDAR feature map aligned to the same BEV grid. FFB projects the joint representation to the dimensions expected by the existing detection head. This feature interface supports the same design across BEVFusion-MIT, BEVFusion-AD, and GraphBEV. Each detector is adapted and trained separately, using a camera-only checkpoint, router/fusion warm-up with modality dropout, and end-to-end fine-tuning.

\begin{figure*}[!t]
  \centering
  \includegraphics[width=0.8\textwidth]{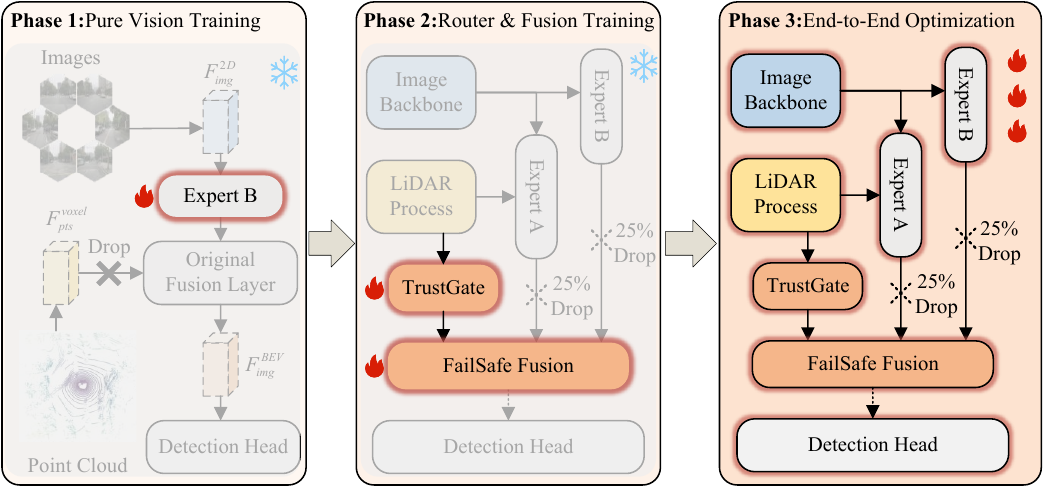}
  \caption{\textbf{Three-Phase Training Strategy.}
  \textbf{Phase 1:} Pure vision pre-training to establish a strong semantic fallback (Expert B).
  \textbf{Phase 2:} Router and Fusion warm-up with MD ($p_{L\text{-}drop}=p_{C\text{-}drop}=0.25$) while freezing the expert backbones.
  \textbf{Phase 3:} End-to-end fine-tuning with the same MD distribution to align the feature spaces of both experts.}
  \label{fig:training_pipeline}
\end{figure*}

\subsection{TrustGate Router: Adaptive Expert Routing}
\label{subsec:trustgate}

The core premise of Grace-BEV is that a fixed mixture of image-to-BEV experts is brittle when the LiDAR signal used by Expert A degrades. The TrustGate Router is a lightweight module that adapts this mixture from LiDAR-side feature statistics.

\noindent\textbf{Mechanism.} As shown in \cref{fig:framework}, the router operates on the LiDAR BEV feature map  $F_{pts}^{voxel} \in \mathbb{R}^{C_{pts} \times H \times W}$, where $C_{pts}$ denotes the channel dimension of the LiDAR backbone.
Instead of employing heavy attention mechanisms, we use global statistics that respond to changes in the LiDAR-guided representation.
More specifically, we employ dual adaptive pooling to aggregate element-wise statistics into a global descriptor $s_{stat} \in \mathbb{R}^{2C}$:
\begin{equation}
    s_{stat} = \text{Concat}\left( \mathcal{P}_{avg}(F_{pts}^{voxel}), \mathcal{P}_{max}(F_{pts}^{voxel}) \right),
    \label{eq:pooling}
\end{equation}
where $\mathcal{P}_{avg}$ and $\mathcal{P}_{max}$ denote global adaptive average and max pooling, respectively.
These statistics capture background activation and salient geometric evidence available to the LiDAR-guided expert.
A lightweight bottleneck MLP maps these statistics to a routing coefficient:
\begin{equation}
    \alpha = \sigma \left( W_2 \cdot \delta ( W_1 \cdot s_{stat} ) \right).
    \label{eq:trust_score}
\end{equation}
Here, $W_1 \in \mathbb{R}^{64 \times 2C}$ maps the input descriptor to a compact hidden representation (fixed at 64 channels), $\delta$ denotes the ReLU activation, and $W_2 \in \mathbb{R}^{1 \times 64}$ maps the hidden features to a scalar.
The output $\alpha$ is reshaped to $\mathbb{R}^{B \times 1 \times 1 \times 1}$ for broadcasting.

\noindent\textbf{Routing Logic.} The router produces a sample-adaptive routing coefficient $\alpha \in [0, 1]$.
We utilize $\alpha$ to softly interpolate the BEV features from the two experts via element-wise broadcasting:
\begin{equation}
    F_{img}^{BEV} = \alpha \odot F_{\text{Expert A}}^{BEV} + (1 - \alpha) \odot F_{\text{Expert B}}^{BEV},
    \label{eq:routing}
\end{equation}
where $F_{\text{Expert A}}^{BEV}$ and $F_{\text{Expert B}}^{BEV}$ denote the projected BEV feature representations generated by Expert A and Expert B, respectively.
Larger $\alpha$ favors LiDAR-guided Expert A, while smaller $\alpha$ favors vision-only Expert B. During joint fine-tuning, this coefficient is optimized together with the expert representations, FFB, and detection head. $\alpha$ is an expert-mixing coefficient optimized for downstream detection, rather than a calibrated estimate of LiDAR sensor reliability or an explicit upper bound on raw LiDAR influence. The raw LiDAR BEV representation enters FFB separately, where an input-dependent element-wise gate recalibrates its interaction with the routed image features.

\subsection{FailSafe Fusion Block}
\label{subsec:failsafe}

Standard fusion approaches typically adopt a ``concatenate-and-convolve'' strategy. While effective under ideal conditions, this hard fusion creates a vulnerability: if the LiDAR backbone output $F_{pts}^{voxel}$ becomes corrupted or zero-filled due to sensor failure, the concatenated tensor undergoes a severe distribution shift, allowing unreliable features to propagate into subsequent layers.

To handle this, we propose the FFB, which introduces a learnable gating mechanism to dynamically recalibrate feature importance before integration.

\noindent\textbf{Architecture.} As illustrated in \cref{fig:framework}, the fusion block receives two inputs: the raw LiDAR BEV features $F_{pts}^{voxel}$ from the LiDAR backbone and the routed image-BEV features $F_{img}^{BEV} \in \mathbb{R}^{C_{img} \times H \times W}$ produced by Eq. \ref{eq:routing}.
We first concatenate them to form a unified representation:
\begin{equation}
    F_{cat} = \text{Concat}(F_{pts}^{voxel}, F_{img}^{BEV}) \in \mathbb{R}^{(C_{pts}+C_{img}) \times H \times W}.
\end{equation}
Instead of fusing them directly, we employ a \textbf{Gate Generator} to predict an element-wise re-weighting map $W_{gate}$ from this joint context:
\begin{equation}
    W_{gate} = \sigma \left( \mathcal{F}_{excite}( \mathcal{F}_{squeeze}(F_{cat}) ) \right).
    \label{eq:gate_gen}
\end{equation}
The generator uses a lightweight bottleneck design (reduction ratio $r=4$).
Specifically, $\mathcal{F}_{excite}$ uses a dilated convolution ($3 \times 3$, $d=2$) to predict recalibration weights from spatial context over an enlarged receptive field.

\noindent\textbf{Gated Integration.} The learned gate $W_{gate}$ acts as a soft mask to filter out noisy or corrupted channels in the concatenated tensor:
\begin{equation}
    F_{fused}^{BEV} = \mathcal{F}_{align} \left( F_{cat} \odot W_{gate} \right),
    \label{eq:gated_fusion}
\end{equation}
where $\odot$ denotes element-wise multiplication.
When the LiDAR stream $F_{pts}^{voxel}$ is corrupted, the learned gate can reduce corresponding elements and retain more of the routed image representation.
Finally, $\mathcal{F}_{align}$ projects the gated features to the target dimension required by the detection head.
By initializing the final bias of the Gate Generator to 0 (yielding $W_{gate} \approx 0.5$), we ensure an unbiased starting point for stable training.

\subsection{Three-Phase Robust Training Strategy}
\label{subsec:training}

Training a robust MoE framework is non-trivial due to the ``modality dominance'' phenomenon~\cite{yan2023cmt,peng2022balanced,wu2022characterizing}, where the model tends to over-rely on the dominant modality (LiDAR) while ignoring the weaker one (Camera).
To address this, we design a \textbf{Three-Phase Training} pipeline, as illustrated in \cref{fig:training_pipeline}.

\noindent\textbf{Phase 1: Pure Vision Pre-training.}
We initialize Expert B's backbone and LSS view transformer from a pre-trained camera-only detector and train this pathway without LiDAR~\cite{liu2022bevfusion,liang2022bevfusion}. This establishes a visual fallback before joint optimization.

\noindent\textbf{Phase 2: Router \& Fusion Warm-up.}
We freeze the image backbone, Expert B, and the LiDAR backbone of Expert A, and train TrustGate and FFB on these fixed representations. With modality dropout, we sample the raw sensor inputs:
\begin{equation}
    (\tilde{X}_{lidar}, \tilde{X}_{img}) = \begin{cases}
    (X_{lidar}, X_{img}), & p=0.50 \\
    (\varnothing, X_{img}), & p=0.25 \\
    (X_{lidar}, \mathbf{0}), & p=0.25
    \end{cases},
    \label{eq:modality_dropout}
\end{equation}
where $\varnothing$ denotes an empty point cloud and $\mathbf{0}$ denotes zero-filled images; feature extraction is then applied to the sampled inputs~\cite{yan2023cmt}.
The LiDAR-failure case encourages greater use of Expert B, while the camera-failure case trains fusion to retain usable raw LiDAR features.

\noindent\textbf{Phase 3: End-to-End Fine-tuning.}
We unfreeze all components and jointly optimize the expert representations, router, fusion module, and detection head with a reduced learning rate. We maintain the same modality-dropout distribution defined in Eq.~\eqref{eq:modality_dropout} during this phase.

\begin{table*}[!t]
\centering
\caption{Quantitative Comparison of Robustness against Modality Failures.
\textbf{RD}: Robustness Density, calculated as $(\text{mAP}_{\text{L-Drop}} + \text{mAP}_{\text{C-Drop}}) / \text{Params}$.
\textbf{MRE}: Marginal Robustness Efficiency, defined as the total robustness gain per unit of extra parameters: $\Delta(\text{mAP}_{\text{L-Drop}} + \text{mAP}_{\text{C-Drop}}) / \Delta\text{Params}$.
We compare our method against Zero-shot baselines and state-of-the-art robust approaches.
\textbf{NDS}: nuScenes Detection Score.}
\label{tab:main_results}
\renewcommand{\arraystretch}{1}
\resizebox{\textwidth}{!}{%
\begin{tabular}{lc *{14}{w{c}{0.88cm}}}
\toprule

\multirow{5}{*}{Method} & \multirow{4}{*}{\makecell{Param\\(M)}} & \multicolumn{2}{c}{\multirow{4}{*}{\makecell{Robustness \\ Efficiency}}} & \multicolumn{2}{c}{\multirow{4}{*}{Clean}} & \multicolumn{8}{c}{LiDAR failures} & \multicolumn{2}{c}{Camera failures} \\
\cmidrule(lr){7-14} \cmidrule(lr){15-16}

 & & \multicolumn{2}{c}{} & \multicolumn{2}{c}{} & \multicolumn{2}{c}{Beam Reduction} & \multicolumn{2}{c}{LiDAR Drop} & \multicolumn{2}{c}{Limited FOV} & \multicolumn{2}{c}{Object Failure} & \multicolumn{2}{c}{View Drop} \\

 & \multirow{1}{*}{} & \multicolumn{2}{c}{} & \multicolumn{2}{c}{} & \multicolumn{2}{c}{\textit{4 beams}} & \multicolumn{2}{c}{\textit{all}} & \multicolumn{2}{c}{\textit{[-60, 60]}} & \multicolumn{2}{c}{\textit{rate = 0.5}} & \multicolumn{2}{c}{\textit{6 drops}} \\
 \cmidrule(lr){3-4} \cmidrule(lr){5-6} \cmidrule(lr){7-8} \cmidrule(lr){9-10} \cmidrule(lr){11-12} \cmidrule(lr){13-14} \cmidrule(lr){15-16}

 & & RD & MRE & mAP & NDS & mAP & NDS & mAP & NDS & mAP & NDS & mAP & NDS & mAP & NDS \\
\midrule

\multicolumn{16}{l}{\textit{\textbf{I. Zero-shot Baselines}}} \\
DETR3D~\cite{wang2022detr3d}            & 15.6 & - & - & 34.3 & 42.8 & - & - & - & - & - & - & - & - & 0.0 & 0.0 \\
CenterPoint~\cite{yin2021center}        & 6.0  & - & - & 56.3 & 63.4 & 19.7 & 28.9 & 0.0 & 0.0 & 13.2 & 24.3 & 23.4 & 30.1 & - & - \\
TransFusion~\cite{bai2022transfusion}       & 36.9 & - & - & 66.8 & 70.2 & 40.8 & 52.7 & 0.1 & 0.0 & 20.1 & 39.2 & 33.7 & 46.6 & 61.1 & 66.4 \\
BEVFusion-MIT~\cite{liu2022bevfusion}      & 40.8 & - & - & 66.9 & 69.7 & 41.2 & 52.4 & 0.0 & 0.0 & 20.2 & 37.6 & 34.3 & 47.8 & 56.2 & 61.5 \\
BEVFusion-AD~\cite{liang2022bevfusion}       & 90.2 & - & - & 68.1 & 70.1 & 43.4 & 53.9 & 0.0 & 0.0 & 21.1 & 39.7 & 35.2 & 49.4 & 57.3 & 62.1 \\
DeepInteraction~\cite{yang2022deepinteraction}   & 57.9 & - & - & 68.8 & 71.3 & 46.5 & 54.9 & 0.0 & 0.0 & 21.4 & 39.9 & 41.7 & 53.2 & 59.9 & 64.0 \\
GraphBEV~\cite{song2024graphbev}          & 41.4 & - & - & 68.9 & 71.3 & 46.1 & 53.4 & 0.0  & 0.0  & 20.9 & 40.6 & 40.6 & 50.5 & 59.2 & 64.2 \\
UniTR~\cite{wang2023unitr}             & 12.2 & - & - & 70.2 & 72.1 & 46.8 & 56.5 & 0.1 & 0.8 & 21.7 & 41.1 & 39.3 & 55.1 & 62.4 & 66.1 \\
SparseFusion~\cite{xie2023sparsefusion}      & 43.8 & - & - & 70.9 & 72.9 & 48.1 & 59.6 & 0.0 & 0.0 & 23.8 & 44.1 & 41.1 & 59.2 & 57.8 & 65.9 \\
\midrule
\multicolumn{16}{l}{\textit{\textbf{II. Existing Robust Approaches}}} \\
\multicolumn{16}{l}{\quad \textit{-- Transformer \& Query-based}} \\
CMT~\cite{yan2023cmt}                 & 86.6     & 1.1 & - & 69.7 & 72.3 & 52.4 & 60.1 & 37.8 & 42.4 & 41.7 & 52.6 & 64.8 & 69.2 & 61.0 & 67.2 \\
MoME~\cite{mome}                  & 86.6+2.5 & 1.1 & 1.3 & 70.3 & 72.1 & 53.4 & 61.6 & 40.6 & 46.1 & 48.6 & 53.3 & 65.8 & 69.4 & 61.4 & 67.7 \\
MEFormer~\cite{cha2024meformer}                        & 86.6+0.3 & 1.2 & 14.3 & 70.5 & 72.7 & 54.3 & 62.1 & 41.0 & 47.3 & 49.9 & 55.2 & 67.1 & 71.8 & 62.1 & 68.3 \\
\addlinespace[0.1cm]
\multicolumn{16}{l}{\quad \textit{-- LSS-based}} \\
BEVFusion-MIT (MD)~\cite{liu2022bevfusion}              & 40.8      & 1.4 & - & 56.1 & 62.2 & 18.1 & 31.3 & 12.1 & 14.4 & 11.8 & 15.2 & 22.5 & 26.6 & 44.1 & 50.1 \\
UniBEV~\cite{wang2024unibev}    & 40.8+31.9 & 1.2 & 1.1 & 63.2 & 66.9 & 45.3 & 52.7 & 33.5 & 40.2 & 38.9 & 48.1 & 57.2 & 64.3 & 57.1 & 63.2 \\
\midrule
\multicolumn{16}{l}{\textit{\textbf{III. Grace-BEV Integrated into LSS-based Detectors}}} \\
\rowcolor{gray!10}
GraphBEV~\cite{song2024graphbev}      & 41.4     & 1.4 & - & 68.9 & 71.3 & 46.1 & 53.4 & 0.0   & 0.0   & 20.9 & 40.6 & 40.6 & 50.5 & 59.2 & 64.2 \\
GraphBEV +\textbf{Ours}         & 41.4+1.2 & 2.3 & 32.7 & 69.3 & 71.8 & 49.2 & 56.3 & 34.2  & 41.7  & 42.8 & 53.1 & 55.2 & 63.6 & 64.2 & 69.1 \\
\textit{improvements}           & -        & - & - & \gain{0.4} & \gain{0.5} & \gain{3.1} & \gain{2.9} & \gain{34.2} & \gain{41.7} & \gain{21.9} & \gain{12.5} & \gain{14.6} & \gain{13.1} & \gain{5.0} & \gain{4.9} \\
\rowcolor{gray!10}
BEVFusion-AD~\cite{liang2022bevfusion}         & 90.2     & 0.6 & -    & 68.1 & 70.1 & 43.4 & 53.9 & 0.0   & 0.0   & 21.1 & 39.7 & 35.2 & 49.4 & 57.3 & 62.1 \\
BEVFusion-AD  +\textbf{Ours}    & 90.2+1.2 & 1.1 & 35.9 & 69.2 & 70.8 & 49.0 & 57.3 & 34.7  & 41.5  & 42.1 & 51.9 & 60.9 & 65.6 & 65.7 & 69.6 \\
\textit{improvements}           & -        & -   & -    & \gain{1.1} & \gain{0.7} & \gain{5.6} & \gain{3.4} & \gain{34.7} & \gain{41.5} & \gain{21.0} & \gain{12.2} & \gain{25.7} & \gain{16.2} & \gain{8.4} & \gain{7.5} \\
\rowcolor{gray!10}
BEVFusion-MIT~\cite{liu2022bevfusion}        & 40.8     & 1.4 & -   & 66.9 & 69.7 & 41.2 & 52.4 & 0.0   & 0.0   & 20.2 & 37.6 & 34.3 & 47.8 & 56.2 & 61.5 \\
BEVFusion-MIT +\textbf{Ours}    & 40.8+1.2 & 2.3 & 33.5 & 68.3 & 71.2 & 46.9 & 58.4 & 32.8  & 39.9  & 41.1 & 50.2 & 59.9 & 65.3 & 63.6 & 68.8 \\
\textit{improvements}           & -        & - & - & \gain{1.4} & \gain{1.5} & \gain{5.7} & \gain{6.0} & \gain{32.8} & \gain{39.9} & \gain{20.9} & \gain{12.6} & \gain{25.6} & \gain{17.5} & \gain{7.4} & \gain{7.3} \\
\bottomrule
\end{tabular}%
}
\end{table*}

\begin{figure*}[!t]
  \centering
  \includegraphics[width=0.88\textwidth]{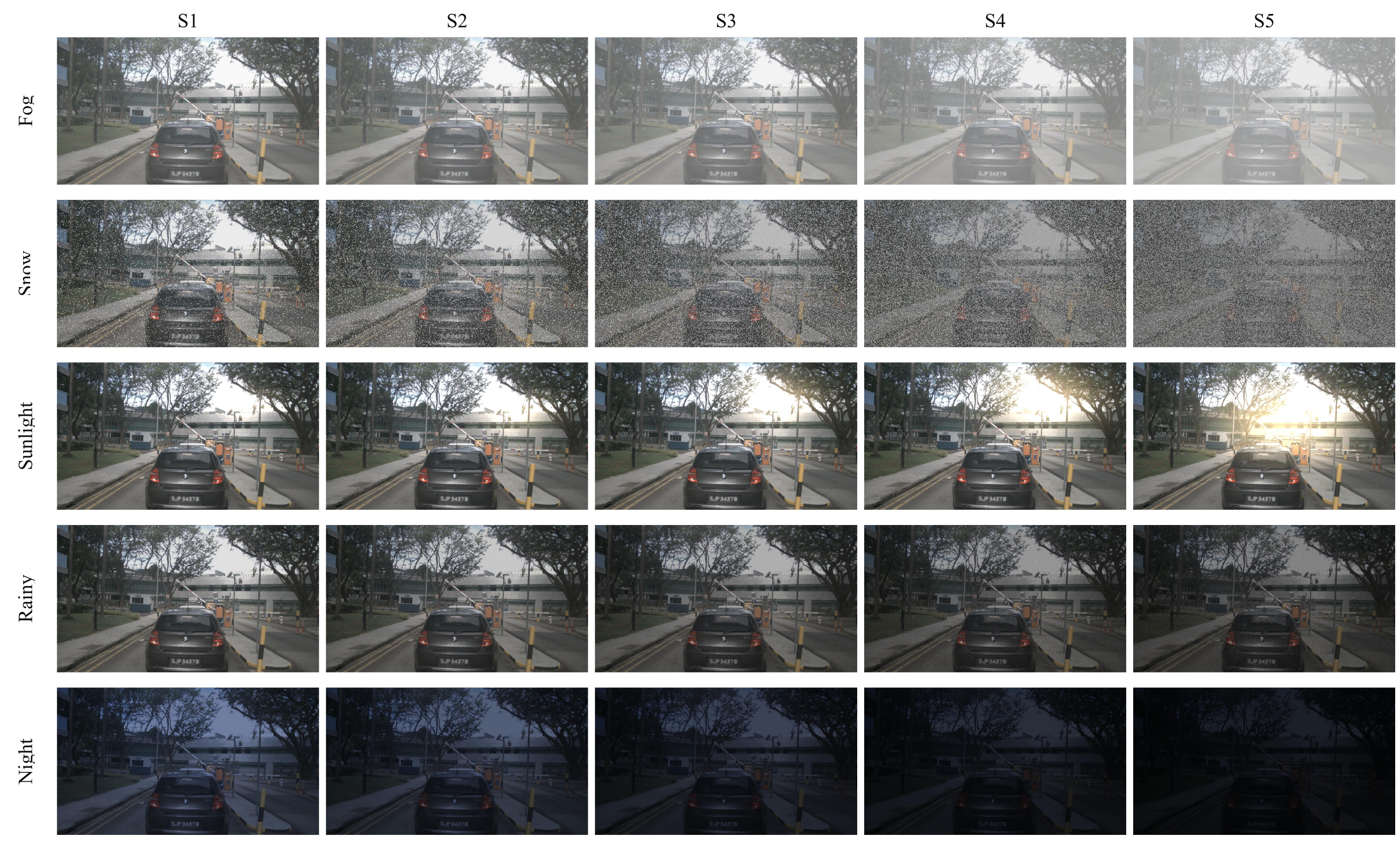}
  \caption{\textbf{Qualitative examples of five-level adverse weather corruptions.}
  Each row shows one weather type, and each column shows one severity level from S1 to S5. The same front camera view is used across severity levels for each weather type to make the degradation trend visually comparable.}
  \label{fig:weather_qualitative}
\par\vspace{3pt}
  \input{tables/main/02_weather.tex}
\end{figure*}

\section{Experiments}
\label{sec:Experiments}

\subsection{Implementation Details}
\label{subsec:implementation}

\noindent\textbf{Dataset and Metrics.}
For evaluation, we report the standard mAP and nuScenes Detection Score (NDS)~\cite{caesar2020nuscenes}.
To rigorously assess robustness, we conduct experiments on two challenging dataset variants: nuScenes-R~\cite{yu2023benchmarking} for sensor impairment simulation and nuScenes-C~\cite{dong2023benchmarking} for extreme weather conditions.
Specifically, on nuScenes-R~\cite{yu2023benchmarking}, we evaluate three categories of corruption:
(1) \textit{LiDAR Failure}, including complete sensor blackout (LiDAR Drop), partial view loss (Limited FOV), and beam reduction;
(2) \textit{Camera Failure}, simulating image blackout (View Drop); and
(3) \textit{Object Failure}, simulating reflection noise.
For weather evaluation, we use the simulated corruption settings provided by nuScenes-C~\cite{dong2023benchmarking}, including fog, snow, sunlight, rain, and night.
For each of the three integrated backbones and its baseline, one fixed checkpoint per configuration is evaluated on the same three independently generated corruption realizations for every weather--severity setting; \cref{tab:weather_robustness} reports their mean mAP.
Our weather protocol includes joint camera--LiDAR corruptions: fog perturbs images and applies range-dependent LiDAR point dropout with intensity attenuation; rain perturbs images and applies random LiDAR point dropout with coordinate jitter; snow further adds false LiDAR points; sunlight introduces image glare and a small fraction of noisy LiDAR points; and night is the only camera-only corruption, leaving LiDAR measurements unchanged.

\noindent\textbf{Architecture and Baselines.}
We select three established detectors as our primary baselines \cite{liang2022bevfusion,liu2022bevfusion,song2024graphbev}.
We re-implement and train all baseline methods in a unified experimental setting to ensure reproducible comparisons under the same robustness protocol.
We retain each baseline's backbone and detection-head designs, introduce the vision-only expert pathway, and replace the fusion stage with \textit{TrustGate Router} and FFB integration. Each resulting detector is trained using the procedure below.

\noindent\textbf{Training Settings.}
All models are trained on 4$\times$ NVIDIA A100 GPUs.
We adopt the proposed Three-Phase Training strategy to ensure stable convergence.
In \textit{Phase 1}, the vision expert is initialized from a pre-trained camera-only checkpoint and warmed up for 1 epoch with a learning rate of $1 \times 10^{-4}$.
In \textit{Phase 2}, we freeze both backbones and exclusively train the router and fusion modules for 3 epochs with a learning rate of $1 \times 10^{-5}$. During this phase, we apply the MD strategy ($p_{both}=0.5$, $p_{L\text{-}drop}=0.25$, $p_{C\text{-}drop}=0.25$) to force the router to learn failure-aware policies.
Finally, in \textit{Phase 3}, we unfreeze all components and fine-tune the entire network end-to-end for another 3 epochs with a reduced learning rate of $1 \times 10^{-7}$~\cite{zhu2019class,lin2017focal}. The same modality-dropout distribution is maintained during Phase 3 to preserve robustness throughout end-to-end fine-tuning.

\subsection{Main Results}
\label{subsec:main_results}
\noindent\textbf{Applicability across LSS-based Architectures.} \Cref{tab:main_results} shows improvements across BEVFusion-MIT, BEVFusion-AD, and GraphBEV under clean inputs and all five sensor-failure settings. Clean mAP increases by 0.4--1.4 points, while complete LiDAR-loss mAP rises from zero for the corresponding baseline models to 32.8--34.7. These consistent gains extend from total input loss to partial LiDAR degradation, supporting the same design across the three LSS-based architectures.

\noindent\textbf{Efficiency.} Each implementation adds 1.2M model parameters; all components are trainable during final joint fine-tuning. On BEVFusion-MIT, latency increases from 40.76 to 40.81 ms on one NVIDIA A100 GPU with batch size 1 and one worker.

\subsection{Robustness against Extreme Weather Conditions}
\label{subsec:weather_robustness}

Across the three architectures, Grace-BEV improves mAP over each corresponding baseline in all 25 weather settings (\cref{tab:weather_robustness})~\cite{dong2023benchmarking,yu2023benchmarking}. Gains range from 1.0 to 7.0 points, covering both joint camera--LiDAR corruptions in fog, rain, and snow and camera-only degradation at night. This pattern complements the sensor-drop experiments: preserving alternative feature paths and adapting their interaction benefits detection as sensor evidence changes continuously. \Cref{fig:weather_qualitative} illustrates the visual changes from S1 to S5.

\subsection{Ablation Studies}
\label{subsec:ablation}

We conduct comprehensive ablations on the nuScenes validation set to verify each proposed component's contribution and justify our design choices, using standard BEVFusion-MIT as the baseline.

\begin{table}[!t]
\centering
\caption{\textbf{Component-wise Ablation Analysis.}
We verify the effectiveness of the TrustGate Router (TG) and FFB.
\textbf{Note:} To ensure a fair comparison, \textbf{all models (including the Baseline)} are trained using the MD strategy.
The baseline row corresponds to ``BEVFusion-MIT (MD)'' in \Cref{tab:main_results}.}
\label{tab:ablation_components}

\renewcommand{\arraystretch}{1}
\setlength{\tabcolsep}{4pt}

\small

\begin{tabular}{c c c c c c c c c}
\toprule
\multirow{2}{*}{Row} & \multicolumn{2}{c}{Components} & \multicolumn{2}{c}{Clean} & \multicolumn{2}{c}{L-Drop} & \multicolumn{2}{c}{C-Drop} \\
\cmidrule(lr){2-3} \cmidrule(lr){4-5} \cmidrule(lr){6-7} \cmidrule(lr){8-9}
 & TG & FFB & mAP & NDS & mAP & NDS & mAP & NDS \\
\midrule
1 & - & - & 56.1 & 62.2 & 12.1 & 14.4 & 44.1 & 50.1 \\
2 & \checkmark & - & 67.2 & 69.9 & 28.5 & 35.2 & 56.0 & 61.3 \\
3 & - & \checkmark & 67.5 & 70.3 & 15.5 & 20.2 & 57.8 & 63.2 \\
4 & \checkmark & \checkmark & \textbf{68.3} & \textbf{71.2} & \textbf{32.8} & \textbf{39.9} & \textbf{63.6} & \textbf{68.8} \\
\bottomrule
\end{tabular}
\end{table}

\noindent\textbf{Impact of Key Components.}
\Cref{tab:ablation_components} compares components under the same MD training protocol. With TrustGate retained, FFB raises LiDAR-drop mAP from 28.5 to 32.8; without TrustGate, it raises 12.1 to 15.5. The full configuration also performs best under clean inputs and camera loss.

\begin{table}[!t]
\centering
\caption{\textbf{Comparison of fusion mechanisms under matched training conditions.}
All variants retain the same TrustGate architecture and initialization. They are trained separately with identical expert initialization, three-phase training schedule, modality-dropout (MD) distribution, training data, random seed, and checkpoint-selection rule. The fusion mechanism is the only experimental variable.}
\label{tab:ablation_fusion}
\renewcommand{\arraystretch}{1}
\small
\setlength{\tabcolsep}{2.8pt}
\begin{tabular*}{\columnwidth}{@{\extracolsep{\fill}}l c c c c c c@{}}
\toprule
\multirow{2}{*}{Fusion Strategy}
& \multicolumn{2}{c}{Clean}
& \multicolumn{2}{c}{L-Drop}
& \multicolumn{2}{c}{C-Drop} \\
\cmidrule(lr){2-3}
\cmidrule(lr){4-5}
\cmidrule(lr){6-7}
& mAP & NDS & mAP & NDS & mAP & NDS \\
\midrule
Concat
& 67.2 & 69.9 & 28.5 & 35.2 & 56.0 & 61.3 \\
Addition
& 66.8 & 69.5 & 27.1 & 34.0 & 57.2 & 62.5 \\
SE-Block
& 67.6 & 70.4 & 29.4 & 36.3 & 59.1 & 64.4 \\
CBAM
& 67.9 & 70.8 & 30.6 & 37.5 & 60.4 & 65.7 \\
\textbf{FFB (Ours)}
& \textbf{68.3} & \textbf{71.2}
& \textbf{32.8} & \textbf{39.9}
& \textbf{63.6} & \textbf{68.8} \\
\bottomrule
\end{tabular*}
\end{table}

\noindent\textbf{Effectiveness of FailSafe Fusion.}
\Cref{tab:ablation_fusion} compares five fusion mechanisms under a matched training protocol, with the TrustGate Router and all other training settings kept consistent across variants. The Concat and FFB configurations correspond to Rows~2 and~4 of \Cref{tab:ablation_components}, respectively. FFB achieves the highest mAP and NDS under clean inputs, complete LiDAR loss, and camera loss. Compared with concatenation, FFB improves mAP by 1.1, 4.3, and 7.6 points under these three conditions, respectively. Compared with CBAM, the strongest competing fusion baseline, FFB achieves additional gains of 0.4, 2.2, and 3.2 mAP points. These results support the effectiveness of input-dependent joint-feature recalibration beyond the shared expert routing mechanism.

\begin{table}[!t]
\centering
\caption{\textbf{Object-intersecting localized LiDAR degradation on the nuScenes validation set.}
We remove points from $K\in\{1,2,3\}$ non-overlapping, axis-aligned $20\,\mathrm{m}\times20\,\mathrm{m}$ BEV blocks within the evaluation range, each containing LiDAR returns and intersecting an annotated object. The last row shows mAP and NDS gains over BEVFusion-MIT.}
\label{tab:local_lidar}
\renewcommand{\arraystretch}{1}
\setlength{\tabcolsep}{3.2pt}
\small
\begin{tabular*}{\columnwidth}{@{\extracolsep{\fill}}l c c c c c c@{}}
\toprule
\multirow{2}{*}{Method} & \multicolumn{2}{c}{$K=1$} & \multicolumn{2}{c}{$K=2$} & \multicolumn{2}{c}{$K=3$} \\
\cmidrule(lr){2-3}\cmidrule(lr){4-5}\cmidrule(lr){6-7}
 & mAP & NDS & mAP & NDS & mAP & NDS \\
\midrule
BEVFusion-MIT & 62.8 & 66.8 & 57.6 & 62.5 & 50.9 & 56.7 \\
\textbf{Grace-BEV} & \textbf{66.7} & \textbf{69.8} & \textbf{63.9} & \textbf{67.4} & \textbf{59.8} & \textbf{64.2} \\
Gain & $3.9\uparrow$ & $3.0\uparrow$ & $6.3\uparrow$ & $4.9\uparrow$ & $8.9\uparrow$ & $7.5\uparrow$ \\
\bottomrule
\end{tabular*}
\end{table}

\begin{table}[!t]
\centering
\caption{\textbf{Ablation of TrustGate Statistics.} Comparison of pooling strategies for routing-coefficient prediction.}
\label{tab:ablation_router}
\renewcommand{\arraystretch}{1}
\small

\begin{tabular}{l c c c}
\toprule
Router Input & Dim & mAP (Clean) & mAP (L-Drop) \\
\midrule
Global AvgPool Only & $C$ & 68.0 & 25.4 \\
Global MaxPool Only & $C$ & 68.1 & 29.8 \\
\textbf{Concat (Avg + Max)} & $2C$ & \textbf{68.3} & \textbf{32.8} \\
\bottomrule
\end{tabular}
\end{table}

\begin{figure*}[!t]
  \centering
  \includegraphics[width=0.8\textwidth]{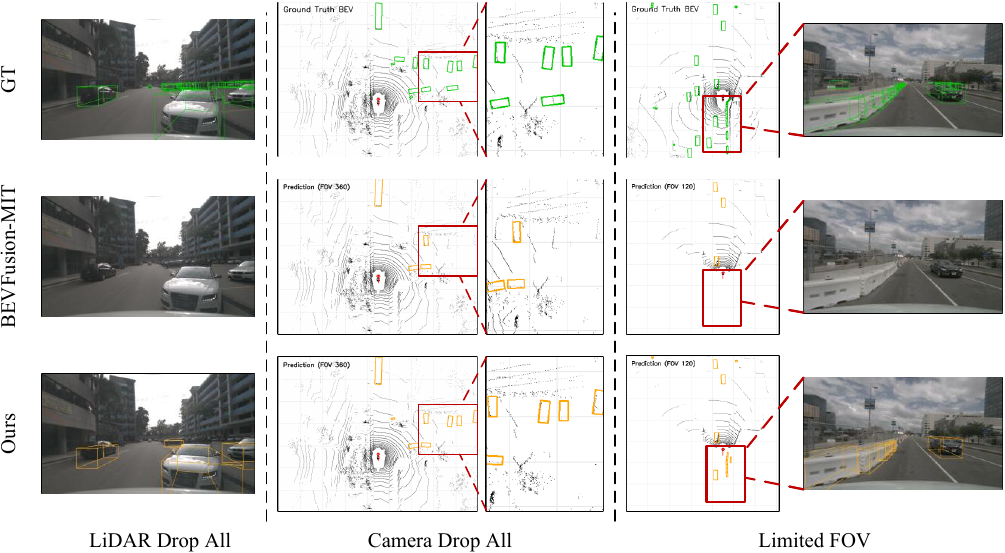}
  \caption{\textbf{Qualitative Comparison under Sensor Corruptions.}
  Visual comparison between Ground Truth (Top), BEVFusion-MIT (Middle), and our Grace-BEV (Bottom) under three sensor-corruption scenarios: complete LiDAR loss, complete camera loss, and limited LiDAR FOV.
  }
  \label{fig:qualitative}
\end{figure*}

\begin{figure*}[!t]
  \centering
  \begin{minipage}[t]{0.30\textwidth}
    \centering
    \includegraphics[width=\linewidth]{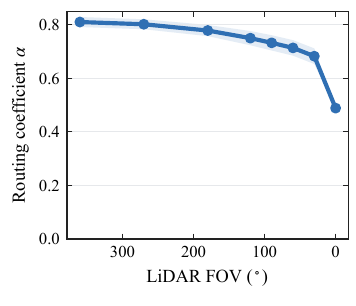}\\[-0.5mm]
    {\footnotesize\textbf{(a) FOV Degradation.}}
  \end{minipage}\hfill
  \begin{minipage}[t]{0.30\textwidth}
    \centering
    \includegraphics[width=\linewidth]{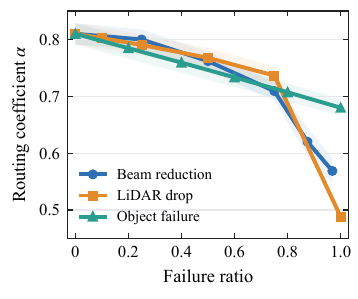}\\[-0.5mm]
    {\footnotesize\textbf{(b) Failure-Ratio Response.}}
  \end{minipage}\hfill
  \begin{minipage}[t]{0.30\textwidth}
    \centering
    \includegraphics[width=\linewidth]{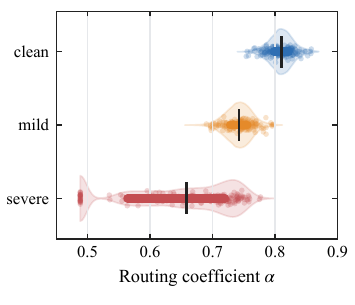}\\[-0.5mm]
    {\footnotesize\textbf{(c) Score Distribution Shift.}}
  \end{minipage}
  \par\vspace{0pt}

  \begin{minipage}[t]{0.34\textwidth}
    \centering
    \includegraphics[width=\linewidth]{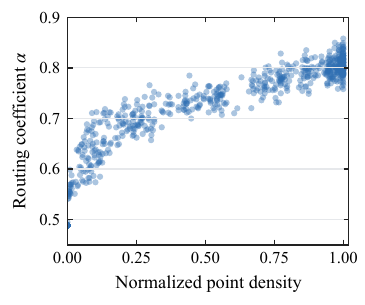}\\[-0.5mm]
    {\footnotesize\textbf{(d) Density Agreement.}}
  \end{minipage}\hspace{0.06\textwidth}
  \begin{minipage}[t]{0.34\textwidth}
    \centering
    \includegraphics[width=\linewidth]{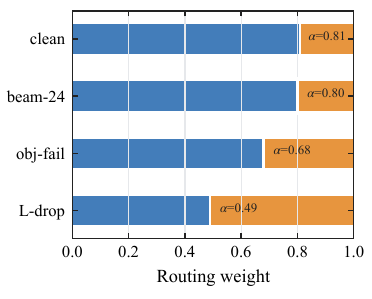}\\[-0.5mm]
    {\footnotesize\textbf{(e) Mean Routing Weights}}
  \end{minipage}
  \caption{\textbf{TrustGate routing under LiDAR degradation.} BEVFusion-MIT + Grace-BEV is evaluated with its checkpoint held fixed across corruption settings.
  \textbf{(a)} Routing coefficient $\alpha$ under LiDAR FOV degradation from $360^{\circ}$ to $0^{\circ}$.
  \textbf{(b)} Routing-coefficient trajectories under beam reduction, LiDAR drop, and object-level failure on a normalized failure-ratio axis; object-level failure uses the measured ratios $0.2/0.4/0.6/0.8/1.0$ without interpolation.
  \textbf{(c)} Sample-level distribution shift of $\alpha$ under clean samples ($s=0$), mild corruptions ($s=1$--$2$), and severe corruptions ($s=4$--$5$).
  \textbf{(d)} Correlation between $\alpha$ and normalized point density.
  \textbf{(e)} Validation-set mean routing weights under representative sensor conditions, illustrating the shift in expert mixing as LiDAR degrades.}
  \label{fig:trustgate_reliability}
\end{figure*}

\noindent\textbf{Router Architecture Design.}
We investigate the input statistics used for the TrustGate Router in \Cref{tab:ablation_router}.
Using only Average Pooling tends to smooth out local high-confidence features, while Max Pooling captures prominent features but may be sensitive to outliers.
Concatenating both statistics allows the MLP to learn a more comprehensive representation of feature integrity, resulting in the highest robustness score.

\begin{table}[!t]
\centering
\caption{\textbf{Ablation on Modality Dropout Probabilities.}
We fix the probability of full-modality training at $p_{both}=0.5$ and vary the ratio of single-modality training.
The default setting ($0.25/0.25$) achieves the best trade-off.
}
\label{tab:ablation_dropout_prob}
\renewcommand{\arraystretch}{1}
\setlength{\tabcolsep}{4pt}

\small

\begin{tabular}{c c c c c c c c}
\toprule
\multicolumn{2}{c}{Dropout Config} & \multicolumn{2}{c}{Clean} & \multicolumn{2}{c}{L-Drop} & \multicolumn{2}{c}{C-Drop} \\
\cmidrule(lr){1-2} \cmidrule(lr){3-4} \cmidrule(lr){5-6} \cmidrule(lr){7-8}
$p_{L\text{-}drop}$ & $p_{C\text{-}drop}$ & mAP & NDS & mAP & NDS & mAP & NDS \\
\midrule
0.00 & 0.50 & 67.8 & 70.5 & 5.2 & 10.5 & 60.8 & 66.4 \\
0.10 & 0.40 & 68.0 & 70.9 & 22.1 & 30.4 & 62.7 & 68.1 \\
\rowcolor{gray!10}
\textbf{0.25} & \textbf{0.25} & \textbf{68.3} & \textbf{71.2} & \textbf{32.8} & \textbf{39.9} & \textbf{63.6} & \textbf{68.8} \\
0.40 & 0.10 & 68.1 & 71.0 & 31.5 & 39.1 & 53.5 & 60.8 \\
0.50 & 0.00 & 67.5 & 70.1 & 32.0 & 39.5 & 34.8 & 46.0 \\
\bottomrule
\end{tabular}
\end{table}

\noindent\textbf{Sensitivity to Dropout Probabilities.}
Finally, we analyze the impact of the MD distribution in \Cref{tab:ablation_dropout_prob}.
We fix the full-modality training probability $p_{both}=0.5$ and vary the allocation of failure simulation.
An imbalanced distribution leads to biased robustness. For instance, setting $p_{L\text{-}drop}=0$ (no LiDAR-drop samples during the modality-dropout training stages) results in a failure to generalize to LiDAR blackouts (5.2 mAP).
Our balanced strategy (0.25/0.25) provides a favorable trade-off between clean accuracy and dual-modality robustness.

\subsection{Spatially Localized LiDAR Degradation}
\label{subsec:local_lidar}

Spatially localized LiDAR loss leaves a mixture of intact and depleted geometry within each frame. \Cref{tab:local_lidar} tests whether Grace-BEV maintains detection accuracy as more object-relevant regions lose their LiDAR returns.

Grace-BEV outperforms BEVFusion-MIT at every degradation level, with the mAP advantage increasing from 3.9 points at $K=1$ to 8.9 points at $K=3$. The complete detector therefore retains a growing advantage as LiDAR evidence becomes spatially fragmented around annotated objects.

\begin{figure*}[!t]
  \centering
  \begin{minipage}[t]{0.47\textwidth}
    \centering
    \includegraphics[width=0.96\linewidth]{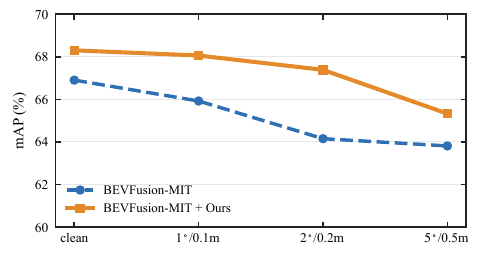}\\[-0.5mm]
    {\footnotesize\textbf{(a) Extrinsic Calibration Noise.}}
  \end{minipage}\hfill
  \begin{minipage}[t]{0.47\textwidth}
    \centering
    \includegraphics[width=0.96\linewidth]{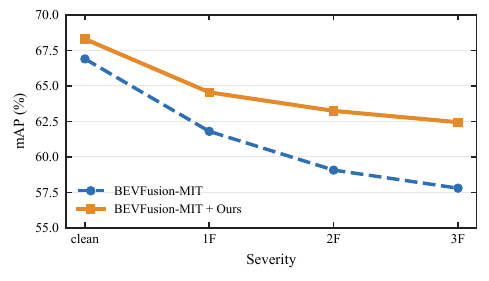}\\[-0.5mm]
    {\footnotesize\textbf{(b) Camera-LiDAR Temporal Delay.}}
  \end{minipage}
  \caption{\textbf{Robustness under deployment-oriented failures.}
  \textbf{(a)} Extrinsic calibration noise with coupled yaw and translation perturbations.
  \textbf{(b)} Camera-LiDAR temporal delay. All values are measured mAP on BEVFusion-MIT; Grace-BEV remains the higher curve under both deployment-oriented failures.}
  \label{fig:deployment_failures}
\end{figure*}

\subsection{Qualitative Analysis}
\label{subsec:qualitative}

\textbf{Qualitative Results.} \cref{fig:qualitative} presents qualitative results comparing Grace-BEV with the baseline BEVFusion-MIT across sensor failure scenarios.
While the baseline is significantly affected by modality failures, Grace-BEV preserves more detections through the combined fallback, routing, and fusion.

\begin{table*}[!t]
\centering
\caption{\textbf{Routing policy comparison under modality corruptions.}
Rows change the image-BEV mixing coefficient in the final trained detector: forced B/A uses $\alpha=0/1$, fixed gating uses $\alpha=0.5$, and random gating provides a stochastic control. Columns apply sensor corruptions to the full detector, including raw LiDAR entering FFB. All policies use the same network parameters, validation frames, and corruption realizations; the input-dependent FFB gate is recomputed from the resulting features. Values are mAP.}
\label{tab:routing_policy}
\renewcommand{\arraystretch}{1}
\setlength{\tabcolsep}{4pt}
\small
\begin{tabular}{l c c c c c c}
\toprule
\multirow{2}{*}{Routing policy} & \multicolumn{5}{c}{Failure setting} & \multirow{2}{*}{Avg.} \\
\cmidrule(lr){2-6}
 & Beam red. & LiDAR drop & Limited FOV & Object fail. & Camera drop & \\
\midrule
Route to Expert B & 32.6 & 31.6 & 38.0 & 43.3 & 49.8 & 39.1 \\
Route to Expert A & 0.1 & 0.0 & 10.2 & 25.3 & 61.2 & 19.4 \\
Fixed gate ($\alpha=0.5$) & 25.6 & 24.1 & 33.7 & 42.6 & 51.6 & 35.5 \\
Random gate & 24.3 & 22.8 & 32.8 & 41.9 & 50.7 & 34.5 \\
Point-count heuristic & 44.8 & 29.1 & 36.6 & 53.8 & 58.9 & 44.6 \\
Point-density heuristic & 45.3 & 29.4 & 37.0 & 54.2 & 59.4 & 45.1 \\
FOV-coverage heuristic & 44.5 & 28.9 & 37.4 & 53.5 & 58.7 & 44.6 \\
\textbf{TrustGate} & \textbf{46.9} & \textbf{32.8} & \textbf{41.1} & \textbf{59.9} & \textbf{63.6} & \textbf{48.9} \\
\bottomrule
\end{tabular}
\par\smallskip
\begin{minipage}{\textwidth}
\footnotesize
\textit{Heuristic definitions.} These availability indicators reflect the density and FOV changes studied in corruption benchmarks~\cite{dong2023benchmarking}. Point count uses valid returns within the evaluation range; density is points per occupied BEV cell; FOV coverage is the fraction of occupied $5^{\circ}$ azimuth bins. Zero-filled entries are excluded, and each statistic is zero for empty inputs. For this control, we map each statistic $z$ to $\alpha_h=\sigma\!\left(4\left[(z-q_{10})/\max(q_{50}-q_{10},\epsilon)-0.5\right]\right)$, where $q_{10}$ and $q_{50}$ are clean-training percentiles and $\epsilon=10^{-6}$. The mapping is fixed before validation and overrides only $\alpha$ in Eq.~\eqref{eq:routing}.
\end{minipage}
\end{table*}

\begin{table}[!t]
\centering
\caption{\textbf{Ablation of Three-Phase Training.}
BEVFusion-MIT + Grace-BEV mAP with clean input or complete loss of one sensor.}
\label{tab:training_strategy_ablation}
\renewcommand{\arraystretch}{1.12}
\small
\begin{tabular*}{\columnwidth}{@{\extracolsep{\fill}}l c c c@{}}
\toprule
Training variant & Clean & \makecell{LiDAR\\drop} & \makecell{Camera\\drop} \\
\midrule
End-to-end training & 41.6 & 4.8 & 38.9 \\
Without Phase 1 & 23.4 & 2.1 & 21.8 \\
Phase 1 from scratch & 64.2 & 17.6 & 55.1 \\
No router/fusion warm-up & 65.1 & 21.4 & 56.8 \\
Unfrozen router/fusion warm-up & 65.6 & 24.7 & 57.4 \\
\textbf{Full three-phase training} & \textbf{68.3} & \textbf{32.8} & \textbf{63.6} \\
\bottomrule
\end{tabular*}
\end{table}

\subsection{Routing Responses to LiDAR Degradation}
\label{subsec:trustgate_reliability}

As degradation increases in \cref{fig:trustgate_reliability}, the coefficient generally decreases, and its distribution shifts downward with substantial overlap. Higher normalized point density is associated with larger $\alpha$.

In \cref{fig:trustgate_reliability}(e), the validation-set mean routing coefficient $\alpha$ decreases from 0.81 under clean inputs to approximately 0.49 under complete LiDAR dropout, indicating a shift in the learned expert mixture toward Expert B. This routing response is learned jointly with the expert representations and downstream fusion during end-to-end fine-tuning.

\subsection{Robustness under Calibration Noise and Temporal Delay}
\label{subsec:deployment_failures}

Beyond zero-filled sensor failures and adverse weather, deployed autonomous systems can suffer from geometric and temporal inconsistencies between cameras and LiDAR. We therefore consider two compact but practical stress tests. The first perturbs the camera-LiDAR extrinsic calibration with coupled yaw and translation noise, using three severity levels: $\pm1^{\circ}$/0.1 m, $\pm2^{\circ}$/0.2 m, and $\pm5^{\circ}$/0.5 m. The second introduces camera-LiDAR temporal misalignment by delaying one modality by 1, 2, or 3 frames. We evaluate BEVFusion-MIT with and without Grace-BEV under the same inference protocol.

\Cref{fig:deployment_failures} shows that Grace-BEV retains higher mAP as calibration noise or temporal delay increases. The advantage under both perturbations supports adaptation to inconsistent cross-modal evidence beyond complete sensor dropout.

\subsection{Routing-Policy Control Analysis}
\label{subsec:routing_policy}

We can see from \Cref{tab:routing_policy} that TrustGate outperforms fixed routing, random routing, single-expert selection, and the three heuristic policies in all five failure settings. The heuristic policies average 44.6--45.1 mAP, compared with 48.9 for TrustGate. Unlike heuristic policies that determine routing from predefined mappings of raw-point statistics, TrustGate predicts its expert-mixing coefficient from global average- and max-pooled LiDAR BEV features and learns it through the downstream detection objective. During end-to-end fine-tuning, the router is optimized jointly with both experts, FFB, and the detection head, adapting the expert mixture to task-relevant feature responses. This advantage across heterogeneous failures supports learning expert mixing from task-trained feature responses.

The standalone Phase~1 camera-only detector, using its own detection head without TrustGate or FFB, achieves 32.6 mAP and 38.2 NDS on the same validation split, establishing a strong visual reference. Under complete LiDAR loss, learned routing in the final model reaches 32.8 mAP and 39.9 NDS, compared with 31.6 mAP and 35.3 NDS for forced Expert B. With FFB and the jointly trained detection head retained, this corresponds to a routing gain of 1.2 mAP and 4.6 NDS points.

Together with the fusion comparison in \Cref{tab:ablation_fusion}, these results support establishing visual capability and then learning its interaction with LiDAR-related representations.

\subsection{Ablation of Three-Phase Training}
\label{subsec:training_strategy_ablation}

\Cref{tab:training_strategy_ablation} tests the role of fallback initialization and staged optimization. Skipping Phase~1 or training the full system end to end severely weakens LiDAR-drop performance.

Training Phase~1 from scratch recovers clean accuracy more readily than LiDAR-drop performance, highlighting the value of camera-only pretraining for the visual fallback. Removing warm-up or updating the backbones during warm-up also reduces robustness. This supports learning routing and fusion on stable expert features before joint fine-tuning, complementing the component comparisons in \Cref{tab:ablation_components}.

\FloatBarrier
\section{Conclusion}
\label{sec:Conclusion}
\label{sec:limitations}
In this paper, we propose Grace-BEV, which establishes a LiDAR-independent visual fallback before learning how to combine its features with LiDAR-related representations through expert routing and element-wise fusion. Three-phase training supports this ordering. Across three LSS-based detectors, the design improves detection performance under clean inputs and all five evaluated sensor-failure settings.

The current framework has limitations that guide its further development. TrustGate applies one sample-level expert mixture across BEV locations, while FFB recalibrates the joint features at each location. The consistent gains under object-intersecting LiDAR degradation motivate spatially adaptive routing that can assign different expert weights to intact and degraded regions. The weather evaluation uses controlled simulated perturbations. Real precipitation, illumination, surface contamination, and sensor processing interact over space and time. Evaluation on synchronized real-weather sequences will examine how the observed gains transfer to these coupled effects. The present implementations also rely on compatible LSS-based BEV interfaces. Extending the framework to architectures beyond LSS and to joint camera--LiDAR failure patterns beyond the evaluated weather corruptions will broaden its applicability. These directions build on the visual fallback and adaptive feature interaction established in this study.

\microtypesetup{protrusion=false,expansion=false}
\bibliographystyle{IEEEtran}
\bibliography{refs}
\end{document}